\documentclass[10pt,doublecolumn,twoside]{IEEEtran}
\IEEEoverridecommandlockouts
\usepackage{cite}
\usepackage{amsmath,amssymb,amsfonts}
\usepackage{algorithmic}
\usepackage{graphicx}
\usepackage{textcomp}
\usepackage{epstopdf}
\usepackage{float}
\usepackage{esint}
\usepackage{subfigure}
\usepackage{color}
\usepackage{mathrsfs}
\usepackage{setspace}

\def\BibTeX{{\rm B\kern-.05em{\sc i\kern-.025em b}\kern-.08em
    T\kern-.1667em\lower.7ex\hbox{E}\kern-.125emX}}

\makeatletter

\newcommand{\Rmnum}[1]{\expandafter\@slowromancap\romannumeral #1@}
\makeatother

\newcommand{\ls}[1]
    {\dimen0=\fontdimen6\the\font
     \lineskip=#1\dimen0
     \advance\lineskip.5\fontdimen5\the\font
     \advance\lineskip-\dimen0
     \lineskiplimit=.9\lineskip
     \baselineskip=\lineskip
     \advance\baselineskip\dimen0
     \normallineskip\lineskip
     \normallineskiplimit\lineskiplimit
     \normalbaselineskip\baselineskip
     \ignorespaces
    }

\begin{document}
\title{Queue-Aware Token Communications: Bridging AI Representation and Wireless Service}

\author{Yuquan Xiao, Qinghe Du, Chaoyang Zhang, Wenchi Cheng, Jingqing Wang, Tao Chen, and Jiangzhou Wang

\thanks{Yuquan Xiao, Qinghe Du, and Chaoyang Zhang are with the School of Information and Communications Engineering, Xi'an Jiaotong University, Xi'an 710049, China.

Wenchi Cheng and Jingqing Wang are with the State Key Laboratory of Integrated Services Networks, Xidian University, Xi'an 710071, China.

Tao Chen is with VTT Technical Research Centre of Finland, 02140 Espoo, Finland.

Jiangzhou Wang is with the National Mobile Communications Research Laboratory, Southeast University, Nanjing 210096, China.
}

\vspace{-20pt}
}
\maketitle

\begin{abstract}
The rapid advancement of foundation models and edge intelligence is transforming wireless networks from data-oriented communication systems into artificial intelligence (AI)-native service platforms. Token communications (TokenCom) has emerged as a promising paradigm that directly transmits task-oriented AI representations. However, existing studies typically assume that generated tokens can be transmitted immediately, overlooking the queueing dynamics arising from the mismatch between token workload generation and time-varying wireless service capability. To address this limitation, we introduce Queue-Aware TokenCom, which explicitly incorporates token queues into the AI service pipeline. We present an end-to-end architecture integrating AI representation generation, token workload management, wireless transmission, and downstream task inference. We then discuss two key design directions: budget-aware token representation optimization and queue-aware token service optimization, with the token budget serving as a critical interface between AI performance and communication efficiency. Moreover, we investigate fixed and adaptive token budget strategies to characterize the fundamental tradeoff between AI performance and communication efficiency in dynamic wireless environments. Finally, we outline research directions toward closed-loop, task-aware, and scalable TokenCom systems.
\end{abstract}


\vspace{-10pt}
\section{Introduction}
The evolution of artificial intelligence (AI) is reshaping the role of future wireless networks from merely transporting information to supporting intelligent services~\cite{you2021towards}. Traditional communication systems are primarily designed for bit-level information transmission, where the fundamental objective is to reliably and efficiently deliver raw data. However, with the rapid development of foundation models, multimodal AI, embodied intelligence, and edge intelligence, wireless networks are increasingly expected to support AI inference tasks with stringent requirements on latency, reliability, and efficiency~\cite{Deniz2023Beyond,Wang2023Wireless}. Different from conventional data transmission, AI tasks usually focus on extracting task-relevant information rather than perfectly recovering the original data. This paradigm shift indicates that future communication systems should move beyond transmitting raw data toward delivering task-oriented AI representations. In this context, tokens, which serve as fundamental information units in modern AI models, are emerging as promising communication objects for AI-native networks.

Recent advances in Transformer-based models have established tokens as a key interface between input data and AI inference~\cite{Chen2026Network}. By mapping raw data into compact token representations, modern AI models can effectively capture semantic and task-relevant features for downstream inference. Therefore, token communications (TokenCom) has been proposed as a new communication paradigm, where task-oriented token representations are directly transmitted instead of complete raw data~\cite{Qiao2025Token,Wei2025Token,jiang2026tokencom}. Existing studies mainly focus on reducing token redundancy and communication overhead while preserving AI task performance~\cite{Erak2026Adaptive,Devoto2026Adaptive}. Representative approaches include token selection and token merging for reducing the number of transmitted tokens, as well as feature compression and quantization for reducing token payloads~\cite{lu2026flexselect,bolya2022token}. These efforts demonstrate the potential of token-based transmission for improving the efficiency of AI services.

\begin{figure*}
  \centering
  \includegraphics[scale=0.7]{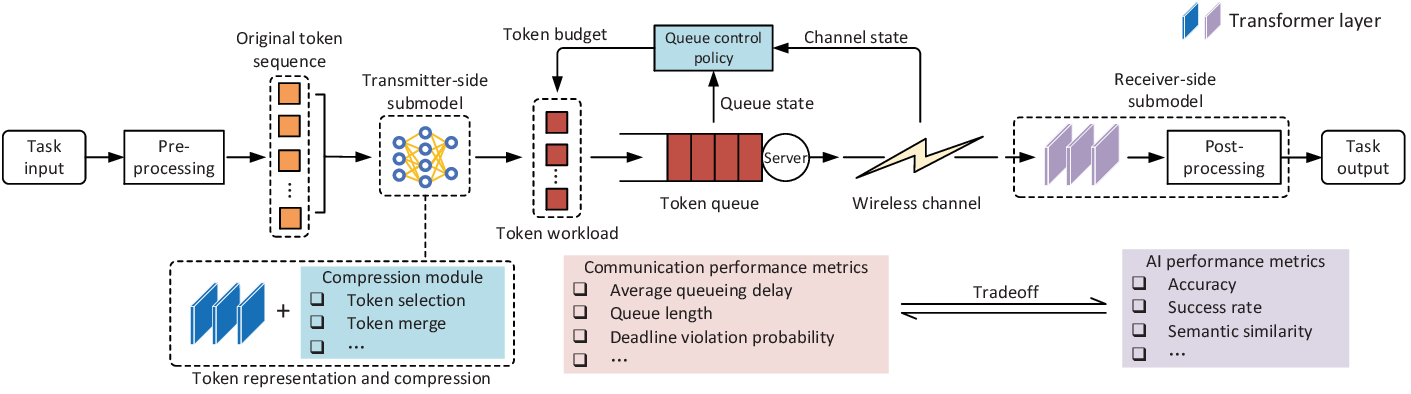}
  \caption{End-to-end architecture of Queue-aware TokenCom. Task-relevant tokens are compressed into a workload, queued, and served over a time-varying wireless channel. The queue-control policy adapts token and wireless decisions according to the queue and channel states, balancing task-level performance with queue-level service.
  \vspace{-10pt}
  }\label{fig:queue_aware_arc}
\end{figure*}

Despite these advantages, existing studies mainly focus on token representation optimization while assuming that generated tokens can be immediately transmitted in a timely manner. Such an assumption may not hold in some practical AI services, where continuous perception, real-time monitoring, and robotic sensing applications continuously generate token workloads~\cite{Xiao2026Age}, while wireless transmission capability dynamically fluctuates due to channel fading, user mobility, and interference. As a result, the token workload arrival process and wireless service capability may become mismatched, leading to accumulated token workloads and additional queueing latency.

To address this limitation, we introduce Queue-aware TokenCom, which jointly considers token representation design and token queue management. By explicitly modeling tokens as communication workloads, Queue-aware TokenCom aims to achieve a better balance between AI task performance and communication service quality under dynamic wireless environments. Following this, the main contribution of this article are summarized as follows: 1) we propose an end-to-end architecture for Queue-aware TokenCom, which establishes a unified pipeline connecting AI representations, token workloads, token queues, and wireless services; 2) we identify the design principles of this emerging paradigm and discuss two fundamental optimization directions: budget-aware token representation optimization and queue-aware token service optimization; 3) we investigate the accuracy-delay tradeoff under fixed and adaptive token budget strategies in dynamic wireless environments; and 4) we highlight several future research directions, including closed-loop TokenCom optimization, task-aware and risk-sensitive TokenCom, and scalable multi-agent TokenCom networks.

The remainder of this article is organized as follows. Section II introduces the architecture, key components, and performance metrics of Queue-aware TokenCom. Section III discusses the design principles and optimization approaches for this paradigm. Section IV presents a case study to demonstrate the effectiveness of Queue-aware TokenCom. Section V highlights potential research directions in this emerging field. Finally, Section VI concludes this article.

\vspace{-10pt}
\section{Queue-aware TokenCom: Architecture, Key Component, and Performance Metrics}
To realize Queue-aware TokenCom, an end-to-end framework is required to integrate AI representation generation, token workload management, and wireless service delivery. In this section, we introduce its overall architecture, key component, and performance metrics.

\vspace{-10pt}
\subsection{End-to-End Architecture}
As illustrated in Fig. 1, the architecture of Queue-aware TokenCom consists of four key components: the transmitter-side submodel, token queue, wireless transmission, and receiver-side submodel. The input data are first preprocessed, e.g., tokenization and embedding, and then transformed by the transmitter-side submodel, in which task-relevant token representations are first extracted for subsequent AI inference. After that, since wireless resources are inherently limited, directly transmitting all generated tokens may introduce excessive communication overhead and lead to severe queueing congestion, and thus a token compression module is employed to adapt the generated token workload by reducing the number, dimension, or precision of transmitted tokens. Note that the representation and compression modules can be also tightly coupled. Various compression strategies, such as token selection and token merging, can be integrated into the transmitter-side submodel to achieve an appropriate balance between task performance and communication efficiency.


The compressed tokens then enter the token queue before wireless transmission. Different from existing TokenCom studies that mainly focus on reducing token redundancy to directly adapt to wireless channel, Queue-aware TokenCom explicitly considers the temporal mismatch between token workload and the transmission capability of wireless channels. In some AI services, token workloads are continuously generated, whereas wireless transmission capability fluctuates due to channel variations, mobility, and interference. Therefore, token queue should be introduced and serves as a critical component for absorbing such mismatches and maintaining the stable AI service delivery.

The wireless transmission module provides services to the token queue according to the available communication resources and channel conditions. Specifically, the token service capability is jointly determined by wireless factors, such as channel quality, bandwidth allocation, transmit power, and scheduling policies. After transmission, the received tokens are processed by the receiver-side submodel, which performs subsequent AI inference and produces the final task outcome. In this way, the complete pipeline of Queue-aware TokenCom is established.

\vspace{-10pt}
\subsection{Token Queue Model}
To characterize the dynamic interaction between token workload and wireless service capability, the token queue evolution can be modeled as
\begin{align}
Q(t+1)=[Q(t)-S(t)]^{+}+A(t),
\end{align}
where \(Q(t)\) denotes the accumulated token workload waiting for transmission, \(A(t)\) represents the arriving token workload generated by AI tasks, and \(S(t)\) denotes the token service capability provided by the wireless transmission module. This queue evolution captures the mismatch between token workload generation and wireless transmission capability. When the arriving token workload exceeds the available service capability, the queue length increases and results in additional queueing latency; otherwise, the accumulated workload can be gradually cleared.

A key feature of the proposed architecture is the Queue-aware control loop. As shown in Fig. 1, the queue control policy observes both queue state and channel state, and dynamically adjusts the token budget or transmission strategy. Specifically, token budget serves as a fundamental control variable that determines how much information can be preserved in the transmitted token representations, while simultaneously affecting the communication workload entering the token queue. Depending on the adopted token compression mechanism, the budget can be characterized from different perspectives, including the number of transmitted tokens, feature dimension, representation precision, or overall payload size.

The impact of token budget on TokenCom can be analyzed from two perspectives. From the AI perspective, token budget directly determines the quality of the received token representations and consequently affects the final inference performance. In general, increasing the token budget enables more task-relevant information to be retained, thereby improving AI task performance. However, the performance gain usually exhibits diminishing returns, since redundant information gradually increases when sufficient semantic information has already been preserved. Therefore, allocating excessive token budget may only provide limited improvement in task performance. From the communication perspective, token budget determines the amount of token workload entering the wireless system. A larger budget results in more transmitted tokens or larger token representations, which increases the workload arrival rate of the token queue. In contrast, a smaller budget reduces the communication burden but may sacrifice the quality of AI representations. Therefore, token budget creates an inherent tradeoff between AI task performance and communication efficiency.

\vspace{-10pt}
\subsection{Performance Evaluation Metrics}
Since the system aims to simultaneously deliver high-quality AI inference and efficient wireless service, two categories of metrics are considered: AI performance metrics that evaluate the quality of AI inference, and communication performance metrics that quantify the efficiency and timeliness of token delivery.
On the one hand, one main objective of TokenCom is to preserve sufficient information for subsequent AI inference. Therefore, AI performance metrics depend on the specific AI applications. For example, image classification tasks can be evaluated by accuracy, object detection tasks can be measured by mean average precision (mAP), while other AI services may adopt task success rate or semantic similarity as performance indicators. These metrics reflect how effectively the transmitted token representations support the target AI tasks. On the other hand, the token queue performance characterizes whether the generated token workload can be efficiently served under dynamic wireless conditions. Among various queue-related metrics, average queueing delay is a critical indicator for real-time AI services, as it directly reflects the waiting time before tokens are delivered for inference. In addition, queue length can be used to characterize workload accumulation, while deadline violation probability can evaluate the reliability of latency-sensitive AI applications.


\vspace{-10pt}
\section{How can We Design and Optimize Queue-aware TokenCom?}
The architecture introduced in the previous section reveals that Queue-aware TokenCom involves a fundamental coupling among AI representations, communication workload, and wireless service. Therefore, the design for this paradigm requires a joint consideration of how much token workload enters the communication system, and how efficiently such workload can be delivered under dynamic wireless conditions. Specifically, the design space can be mainly categorized into two complementary directions: \emph{Budget-aware Token Representation Optimization} and \emph{Queue-aware Token Service Optimization}. The former focuses on generating and compressing task-oriented token representations, aiming to maximize AI task utility given a limited token budget. The latter focuses on efficiently serving the generated token workload by exploiting queue states and wireless resource dynamics, aiming to improve communication efficiency while maintaining satisfactory AI task performance.

\vspace{-10pt}
\subsection{Budget-aware Token Representation Optimization}
Token representation optimization aims to maximize AI task utility under communication budget constraints. Since token budget determines the amount of information that can be delivered through wireless networks, two key design problems need to be addressed: how to compress token representations into a given budget and how to train AI models to adapt to different budget conditions.

\subsubsection{Token-to-Budget Compression}
Given a target token budget, the objective of token-to-budget compression is to reduce the token workload while preserving sufficient information for downstream AI inference. Various compression strategies can be adopted to achieve token-to-budget adaptation. One category is token number reduction, including token selection and token merging. Token selection removes tokens with limited contribution to the target task, while token merging aggregates redundant tokens into compact representations. Another category is token payload reduction, where the feature dimension or representation precision of each token is reduced through feature compression or quantization. Although these approaches differ in implementation, they share the same objective: reducing communication overhead under a given token budget while maximizing the retained task utility.

\begin{figure}
  \centering
  \includegraphics[scale=0.86]{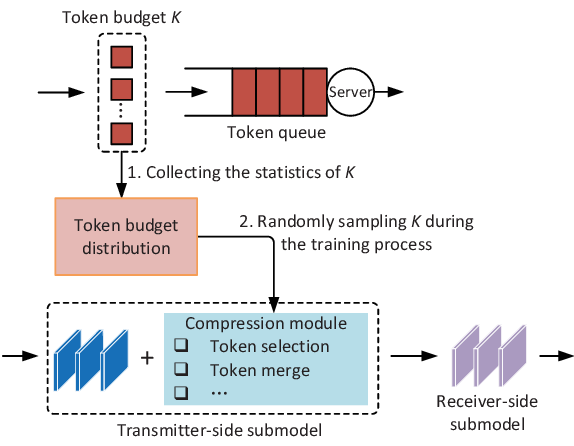}
  \caption{Budget-aware model training. The AI model should be trained by sampling budget values based on the token budget distribution.}\label{fig:model_train}
\end{figure}

\subsubsection{Budget-aware TokenCom Learning}

The compression strategy determines the transmitted token workload, while the AI model determines how effectively such workload can support the target task. Therefore, the transmitter-side submodel, including the compression module, and the receiver-side submodel should be jointly optimized in an end-to-end manner under the target token budget conditions. As illustrated in Fig. 2, the token budget distribution plays a critical role in training budget-aware TokenCom models. During training, the compression module generates token representations according to the available budget, and the complete AI inference pipeline is optimized to maximize task performance under such budget constraints. For fixed-budget systems, the model can be directly trained under the predefined budget. For systems with multiple operating budgets, the budget can be incorporated as a training condition, where different budgets are sampled or weighted according to the target budget distribution.

The matching between training and deployment budget distributions is particularly important. When the budget distribution used during training is inconsistent with the actual deployment distribution, the learned model may not achieve optimal performance in frequently operated budget regions~\cite{fuchs2021distribution}. Therefore, budget-aware TokenCom learning should consider not only end-to-end model optimization but also the statistical characteristics of token budget usage. By aligning model training with practical budget distributions, the TokenCom model can effectively adapt to dynamic communication requirements while maintaining satisfactory AI task performance.

\begin{figure}
  \centering
  \includegraphics[scale=0.9]{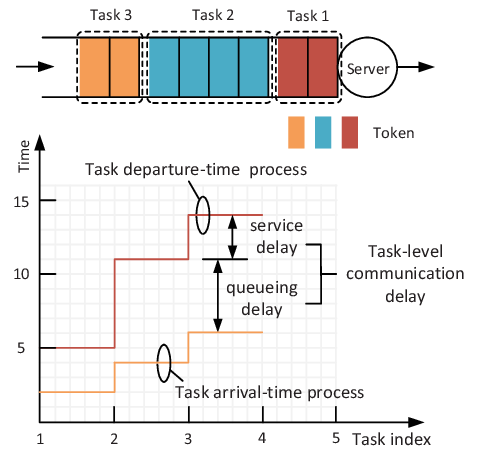}
  \caption{Task-indexed queueing process. Due to the adaptive token budget control, the amount of data to be transmitted varies across tasks. The commonly-used time-indexed queueing process might be challenging to directly characterize the task-level communication delay.}\label{fig:task_index}
\end{figure}

\vspace{-10pt}
\subsection{Queue-aware Token Service Optimization}
While token representation optimization determines the amount and quality of token workload entering the communication system, queue-aware token service optimization focuses on how to efficiently deliver such workload under dynamic wireless conditions. The evolution of the token queue is jointly determined by the token workload arrival process and the wireless service process. Specifically, the token arrival process is affected by multiple factors, including task arrival patterns, token budget allocation, and adopted compression strategies. In contrast, the token service capability is determined by the wireless transmission environment and resource allocation strategies. The mismatch between token workload arrival and wireless service capability leads to dynamic queue evolution, making queue-aware control essential for maintaining efficient AI service delivery. To efficiently serve token workloads, various measures can be conducted by exploiting both AI task characteristics and wireless network dynamics from three aspects:

\subsubsection{Adaptive Token Budget Control}
Adaptive token budget control provides a direct mechanism to regulate the workload entering the queue. By exploiting real-time network states, such as queue length and channel quality, the token budget or compression level can be dynamically adjusted. When the queue pressure increases, more aggressive token compression can reduce workload arrival, while better communication conditions can allow richer token representations to improve AI task performance. Such adaptive control enables a dynamic balance between representation quality and queueing performance.

\subsubsection{Token Scheduling Priority}
Token scheduling determines the service priority among queued tokens. Tokens may have different contributions to the final AI task, and thus scheduling policies can consider semantic importance, task urgency, or inference deadlines to prioritize more valuable tokens and improve overall task performance under limited communication resources.

\begin{figure*}
  \centering
  \includegraphics[scale=0.6]{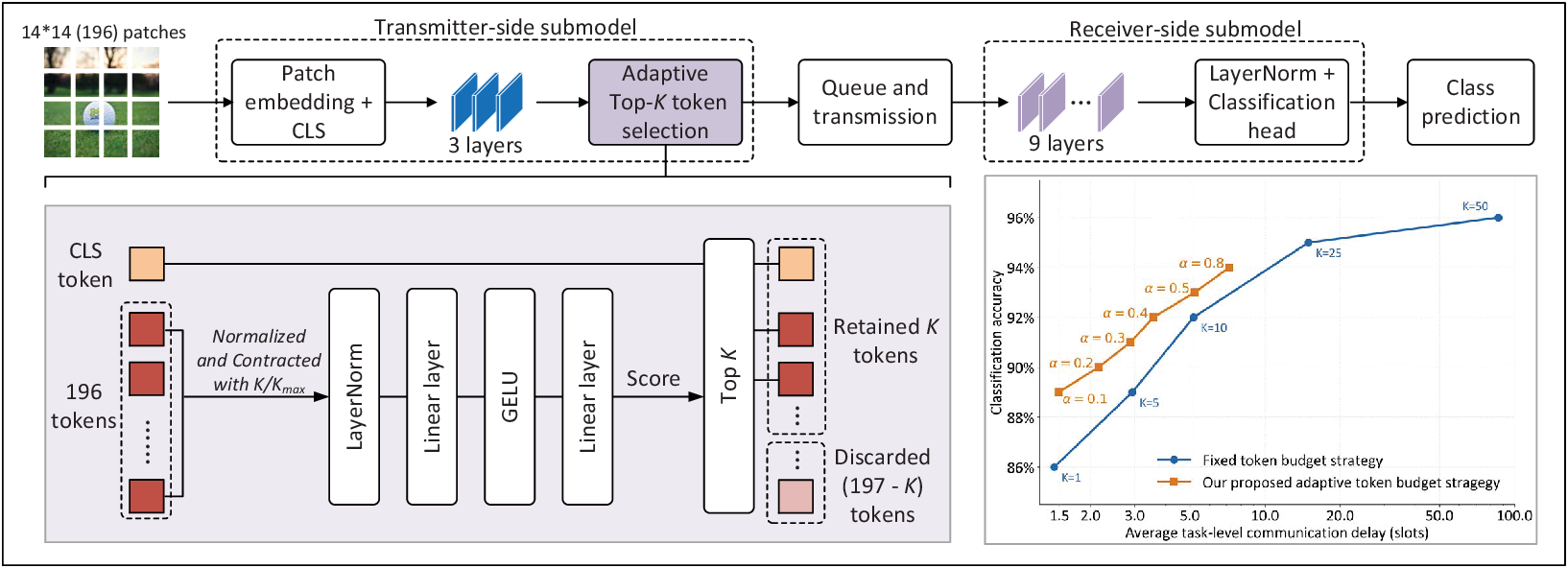}
  \caption{An image classification scenario, where the adaptive Top-$K$ token selection module selects the most informative tokens under a give token budget, and the selected are delivered to the receiver for subsequent inference.
  \vspace{-10pt}}\label{fig:case}
\end{figure*}

\subsubsection{Wireless Resource Allocation}
Wireless resource allocation optimizes the transmission capability for token delivery. By jointly considering queue states and channel conditions, communication resources, such as bandwidth and transmit power, can be dynamically allocated to balance queueing delay and transmission efficiency~\cite{Xiao2023Model}. For example, more resources can be assigned when the queue accumulates significant token workload or when urgent AI tasks require timely delivery.

Note that due to the adaptive token budget control, the amount of data to be transmitted varies across tasks. Then, it becomes challenging to characterize the task-level communication delay based on a time-indexed queueing process, since the relationship between the token backlog and the number of unfinished tasks is no longer straightforward. Therefore, as illustrated in Fig.~\ref{fig:task_index}, a task-indexed modeling approach can be adopted to characterize the communication delay of individual tasks and facilitate its minimization through wireless resource allocation optimization. The entire optimization processes reveal that token representation and queue service are inherently coupled rather than independent. On the one hand, the queue service policy determines the required token budget and compression behavior, thereby influencing the budget distribution experienced during deployment. On the other hand, the resulting AI task performance further affects the optimal communication strategy, since different task requirements may lead to different workload generation and service preferences. Therefore, Queue-aware TokenCom should be viewed as a bidirectional AI-communication system, where token representation optimization and Queue-aware service optimization continuously interact with each other.

\vspace{-10pt}
\section{Case Study: Fixed Token Budget V.S. Adaptive Token Budget}

To validate the effectiveness of Queue-aware TokenCom, we consider an image classification scenario, where intermediate AI representations are treated as communication workloads and delivered through a wireless network before completing the classification task. The adopted model is the DeiT-Tiny/16 vision transformer and the dataset is Imagenette~\cite{Training2021Touvron}. As illustrated in Fig. 3, an input image is first divided into 196 patches, which are then processed by the transmitter-side submodel. Specifically, the 196 patches are embedded into 196 tokens via patch embedding. To collect the global features of the image, a CLS token is also included. Next, these 197 tokens are processed by the three transformer layers of DeiT, and then fed into an adaptive Top-$K$ token selection module, which selects $K$ tokens from the 197 tokens. The structure of the Top-$K$ token selection module is described in the following subsection. The selected tokens by this module then enter a token queue and are transmitted over a wireless channel. The average task queueing delay is used to measure from task arrival until the departure of its last token. Moreover, to characterize the impact of dynamic wireless service capability, we consider a time-correlated Rayleigh fading channel modeled by a first-order Gauss-Markov process, i.e., $h(t)=\rho h(t-1)+\sqrt{1-\rho^2}w(t)$, where $h(t)$ denotes the channel state at time slot $t$, $w(t)$ follows a complex Gaussian distribution, and $\rho$ is the channel correlation coefficient. After receiving the transmitted tokens, the receiver-side submodel, consisting of the remaining nine transformer layers of DeiT and classification head, performs subsequent inference and produces the final classification result.

\vspace{-10pt}
\subsection{Learnable Top-$K$ Token Selection Module}
The token selection module aims to identify the most informative tokens under a given token budget. Instead of using manually designed token importance criteria, we introduce a lightweight learnable scoring network to adaptively evaluate the contribution of each patch token.
As shown in Fig.~\ref{fig:case}, after the third transformer layer, each patch token is first normalized and concatenated with the normalized token budget \(K/K_{\rm max}\), where $K_{\rm max} = 197$. The resulting feature is fed into a shared shallow multi-layer perceptron network, which outputs an importance score for each patch token. Since all tokens share the same scoring network, the module can be efficiently implemented through parallel computation while introducing only a negligible number of additional parameters. Since the CLS token is always preserved, the adaptive Top-$K$ module selects \(K-1\) patch tokens from the 196 patch tokens. Given the obtained scores, the \(K-1\) patch tokens with the highest scores are selected together with the CLS token. Note that the hard Top-$K$ operation is inherently non-differentiable and prevents direct gradient propagation. To enable end-to-end training, we adopt a NeuralSort-style continuous approximation combined with a straight-through estimator. Specifically, the forward pass uses the exact hard Top-$K$ selection to guarantee the desired transmission budget, while the backward pass employs a soft permutation surrogate to propagate gradients through the token scoring network. Therefore, the entire transmitter, token selection module, and receiver can be jointly optimized using the final classification loss.

\vspace{-10pt}
\subsection{Fixed and Adaptive Token Budget Strategies}
As discussed previously, the token budget becomes a critical control variable: a larger \(K\) preserves more task-relevant information but introduces heavier communication workload and potentially larger queueing latency. We consider two token budget strategies to investigate the impact of workload adaptation.

\subsubsection{Fixed Token Budget Strategy}
With this strategy, each task is transmitted using a predefined number of tokens. Such a strategy provides a stable operating point and allows the AI model to be specifically optimized for a given token budget. In addition, this strategy ensures that the incoming data volume remains constant for each image classification task, thereby simplifying the task-level queueing latency modeling. However, it cannot exploit temporal variations of wireless channels. Consequently, when the selected token workload approaches or exceeds the average service capability, accumulated tokens may cause rapidly increasing task completion delay.

\subsubsection{Adaptive Token Budget Strategy}
The key idea of this strategy is to exploit channel predictability and dynamically match token workload with future service capability. At each task arrival instant, the controller observes the current token service capability \(\mu(t)\) and the queue length \(Q(t)\). Here, we propose an adaptive token budget strategy for highly time-correlated wireless channels, where the slot duration is set to 10 ms and the channel correlation coefficient is set to $\rho=0.99$. This setting is reasonable for the uplink transmission of a fixed-position camera, where the channel generally exhibits high temporal correlation at such a short time scale. The arrival interval of image tasks is set to $M = 10$ time slots. Since the channel is highly correlated, the future service capability within two adjacent arrival intervals can be approximately predicted according to the current channel condition. Therefore, the available service opportunity before the next task arrival is estimated as $\hat{C}(t)=M\mu(t)$. The adaptive token budget is then determined as $K_{\rm raw}(t)= \max\{1,\alpha(\hat{C}(t)-Q(t))$, where \(\alpha\) controls the aggressiveness of workload injection. A larger value of \(\alpha\) allows richer token representations but introduces higher queueing risk, while a smaller value of \(\alpha\) adopts a more conservative strategy. Finally, \(K_{\rm raw}(t)\) is projected onto the feasible budget set $K_{\rm set}=\{1,5,10,25,50,100\}$. This adaptive strategy enables the system to allocate more tokens under favorable wireless conditions and reduce token workload during potential congestion periods. Accordingly, the AI model should be trained based on the token budget distribution generated by this adaptive strategy.

To summarize, fixed token budget strategies provide simplicity and predictable behavior, and they allow the AI model to be specifically optimized for a single operating point without parameter sharing among different budgets. In contrast, adaptive token budget strategies require the model to support dynamically varying token budgets, which introduces additional training complexity and potential performance interference due to shared model parameters. However, adaptive strategy provides greater flexibility under dynamic wireless environments by jointly considering channel conditions and queue states.


\begin{figure}
  \centering
  \includegraphics[scale=0.32]{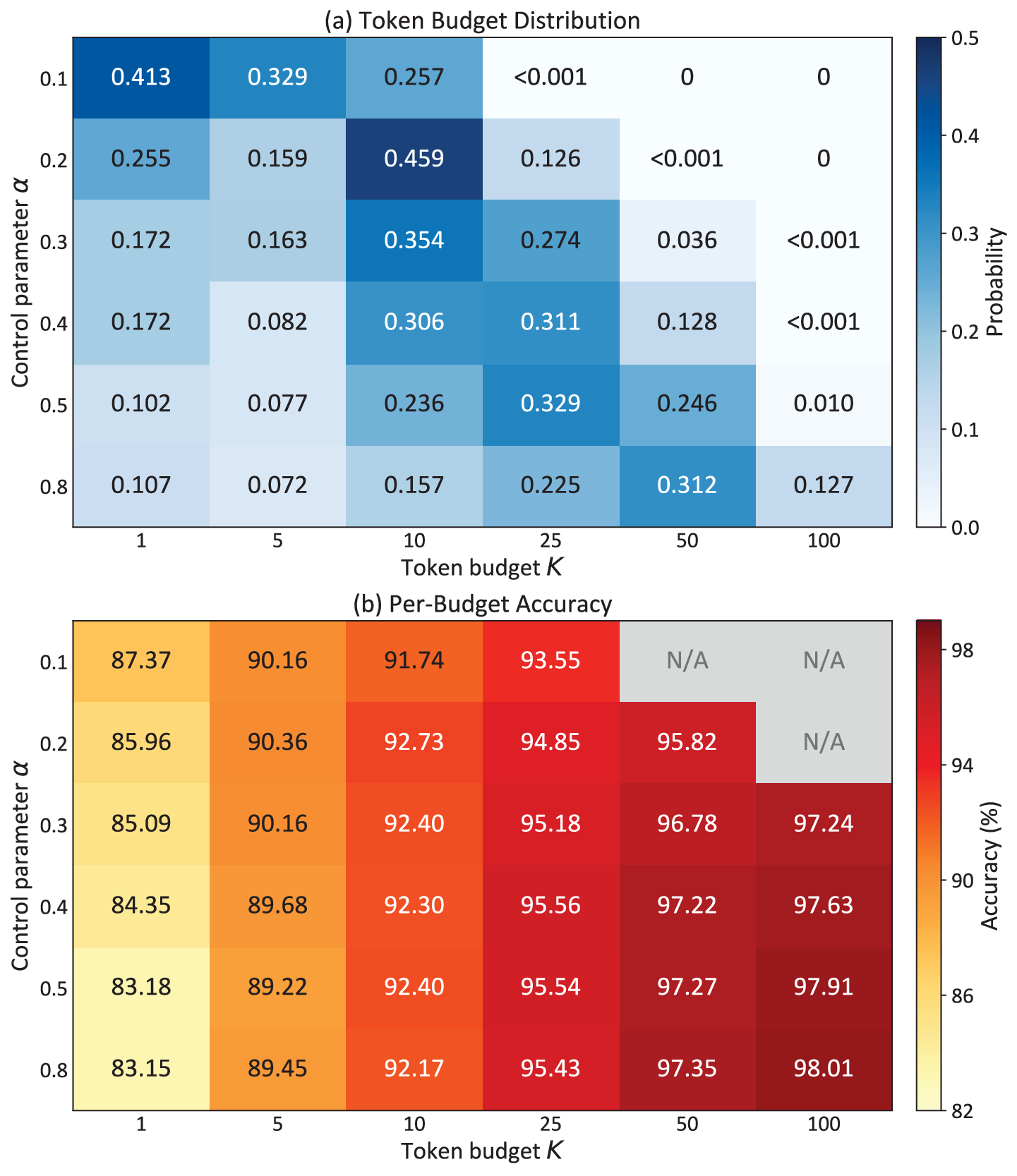}
  \caption{Token budget distributions and per-budget classification accuracy under different values of the control parameter $\alpha$. The results show that changes in the training budget distribution lead to a redistribution of performance across the shared token-budget configurations.
  \vspace{-15pt}}\label{fig:heatmap}
\end{figure}

\vspace{-10pt}
\subsection{Tradeoff Between Classification Accuracy and Task-Level Communication Delay}
We first investigate the fundamental tradeoff introduced by the fixed token budget strategy. As shown in Fig.~\ref{fig:case}, increasing \(K\) consistently improves classification accuracy because more task-relevant visual information is preserved. However, the improvement comes at the cost of significantly increased task-level communication delay due to additional token workload. For example, increasing \(K\) from 25 to 50 only improves accuracy from 95\% to 96\%, while the average task-level communication delay increases from 14.94 slots to 86.06 slots, corresponding to approximately a 5.76-fold increase. This observation indicates that pursuing marginal AI performance improvement with excessive token budget may trigger severe queueing penalties, especially when the workload approaches the wireless service boundary. Next, we evaluate the proposed adaptive token budget strategy. By exploiting the temporal correlation of wireless service capability, the adaptive policy dynamically adjusts the token workload according to the predicted available capacity and queue state. As shown in Fig.~\ref{fig:case}, different values of \(\alpha\) provide flexible operating points between accuracy and delay. Compared with fixed-budget strategies, the adaptive policy achieves a better accuracy-delay trade-off in the low-to-moderate delay regime. For example, at approximately 89\% classification accuracy, the adaptive strategy with \(\alpha=0.1\) reduces the average task-level communication delay from 2.94 slots (fixed \(K=5\)) to 1.49 slots. Similarly, around 92\% accuracy, the adaptive strategy with \(\alpha=0.4\) achieves a delay reduction from 5.17 slots (fixed \(K=10\)) to 3.57 slots. These results demonstrate that adaptive workload control can effectively alleviate unnecessary queue accumulation while maintaining competitive AI performance.

Moreover, as presented in Fig.~\ref{fig:heatmap}, we visualize the token budget distribution and the corresponding per-budget classification accuracy of our proposed adaptive token budget strategy. It can be observed that as \(\alpha\) increases, the budget distribution progressively shifts from smaller to larger token budgets. For example, the sampling probability of \(K=1\) decreases from 0.413 at \(\alpha=0.1\) to approximately 0.107 at \(\alpha=0.8\), while the corresponding accuracy decreases from 87.37\% to 83.15\%. In contrast, larger budgets such as \(K=50\) and \(K=100\) are sampled more frequently under larger values of \(\alpha\) and generally achieve improved accuracy. The results reveal a tendency that the performance associated with a given token budget is positively related to its occurrence frequency during training. This behavior can be explained by the shared-parameter training mechanism of the AI model. The frequently sampled budget contributes to parameter updates more often and receives a larger expected optimization weight. Since all token budgets share the same Transformer backbone and token-scoring network, the model tends to allocate more of its finite optimization capacity to frequently sampled budgets, potentially reducing the specialized performance of rarely sampled budgets.


\vspace{-10pt}
\section{Potential Research Directions}
Although Queue-aware TokenCom provides a new perspective for integrating AI representation and wireless service, several fundamental challenges remain open. In this article, we mainly focus on a single-user scenario and characterize the tradeoff between AI task performance and queue-level service efficiency. However, future AI-native networks are expected to involve dynamic interactions, heterogeneous AI tasks, and multiple intelligent agents. These emerging requirements motivate further investigation into more intelligent, adaptive, and scalable Queue-aware TokenCom systems.

\vspace{-6pt}
\subsection{Closed-Loop TokenCom Optimization}
Queue-aware TokenCom discussed in this article mainly follows a sequential design paradigm, where communication policies determine token budgets and AI models are subsequently optimized to adapt to the resulting token distributions. However, token representation quality and communication service capability are inherently coupled: the optimal token budget depends on AI task requirements and queue states, while the generated token workload further affects future communication decisions. Therefore, future research should investigate closed-loop TokenCom optimization, where AI models, token generation strategies, and queue control policies can continuously interact and adapt according to dynamic task requirements and wireless environments. Such a framework can enable jointly optimized token representation and communication strategies, rather than independently designing them.

\vspace{-6pt}
\subsection{Task-aware and Risk-Sensitive TokenCom}
Future AI services would exhibit highly heterogeneous requirements. For example, autonomous systems and safety-critical applications require not only low latency but also stringent reliability guarantees on task completion. Therefore, future research should investigate task-aware and risk-sensitive TokenCom, where token generation, budget allocation, and queue management are jointly optimized according to the specific requirements of different AI tasks. In particular, new performance metrics beyond average latency, such as deadline violation probability and task failure probability, should be incorporated to support reliable AI services.

\vspace{-6pt}
\subsection{Scalable Multi-Agent TokenCom Networks}
As an emerging AI-native communication paradigm, Queue-aware TokenCom is currently formulated mainly for single-user scenarios, where the token workload generation and queue dynamics are relatively easier to characterize. However, future AI-native networks are expected to support a large number of intelligent agents, such as autonomous robots, vehicles, and embodied AI systems, which collaboratively perform complex tasks through wireless networks. In such multi-agent scenarios, each agent may generate heterogeneous token workloads with different semantic importance, task objectives, and latency requirements. Therefore, scalable Multi-Agent TokenCom requires new task-aware communication mechanisms beyond conventional multi-user scheduling. Future research should investigate intelligent token scheduling, cooperative queue management, and resource allocation strategies that jointly consider task utility, semantic importance, and latency requirements. Moreover, exploiting semantic relationships among agents to enable collaborative token exchange and distributed AI inference represents an important direction toward future AI-native networks.

\section{Conclusion}
In this article, we introduced Queue-aware TokenCom, a new communication paradigm that treats AI tokens as communication workloads and jointly considers token representation, workload management, and wireless service delivery. By incorporating token queues into the AI communication pipeline, the proposed perspective reveals the intrinsic coupling between AI task performance and communication efficiency, where token budget serves as a critical interface connecting AI models and wireless systems. We further discussed key design principles, including token representation optimization and Queue-aware token service optimization, and highlighted future research directions toward closed-loop TokenCom optimization, task-aware and risk-sensitive AI services, and scalable multi-agent TokenCom networks. We envision that Queue-aware TokenCom will provide a promising framework for future AI-native networks, where communication systems can intelligently manage AI workloads by jointly optimizing token generation, queue dynamics, and wireless service delivery.

\vspace{-10pt}

\bibliographystyle{IEEEtran}
\bibliography{References}

\end{document}